# 'The Enemy Among Us': Detecting Hate Speech with Threats Based Othering Language Embeddings

WAFA ALORAINY, Cardiff University, UK
PETE BURNAP, Cardiff University, UK
HAN LIU, Cardiff University, UK
MATTHEW L. WILLIAMS, Cardiff University, UK

Offensive or antagonistic language targeted at individuals and social groups based on their personal character- istics (also known as cyber hate speech or cyberhate) has been frequently posted and widely circulated via the World Wide Web. This can be considered as a key risk factor for individual and societal tension linked to regional instability. Automated Web-based cyberhate detection is important for observing and understanding community and regional societal tension - especially in online social networks where posts can be rapidly and widely viewed and disseminated. While previous work has involved using lexicons, bags-of-words or probabilistic language parsing approaches, they often suffer from a similar issue which is that cyberhate can be subtle and indirect - thus depending on the occurrence of individual words or phrases can lead to a significant number of false negatives, providing inaccurate representation of the trends in cyberhate. This problem motivated us to challenge thinking around the representation of subtle language use, such as references to perceived threats from 'the other' including immigration or job prosperity in a hateful context. We propose a novel framework that utilises language use around the concept of 'othering' and intergroup threat theory to identify these subtleties and we implement a novel classification method using embedding learning to compute semantic distances between parts of speech considered to be part of an 'othering' narrative. To validate our approach we conduct several experiments on different types of cyberhate, namely religion, disability, race and sexual orientation, with F-measure scores for classifying hateful instances obtained through applying our model of 0.93, 0.86, 0.97 and 0.98 respectively, providing a significant improvement in classifier accuracy over the state-of-the-art.

Authors' addresses: Wafa Alorainy, Cardiff University, School of Computer Science and Informatics , Cardiff, Wales, CF24 3AA, UK; Pete Burnap, Cardiff University, School of Computer Science and Informatics , Cardiff, Wales, CF24 3AA, UK; Han Liu, Cardiff University, School of Computer Science and Informatics , Cardiff, CF24 3AA, UK; Matthew L. Williams, Cardiff University, School of Social Sciences, Cardiff, Wales, CF10 3WT, UK.

## 1 INTRODUCTION

As people increasingly communicate through Web-enabled applications, the need for high-quality, automated abusive language detection has become much more necessary. While the benefits of on-line social media are enabling distributed societies to be connected, one unanticipated disadvantage of the technology is the ability for hateful and antagonistic content, or cyberhate, to be published and propagated [9, 51]. Several studies have shown how individuals with biased or negative views towards a range of minority groups are taking to the Web to spread such hateful messages [28, 38]. Instances of cyberhate and racist tension on social media have also been shown to be triggered by antecedent events, such as terrorist acts [8,52].

Expressing discriminatory opinions employs different language uses. For example, to express emotions, hateful words might be used such as *'hate them'*; moreover, to encourage violence, an inflammatory verb could be used such as 'kill'. While these examples contain directly threatening or offensive words *(kill, hate)*, some examples contain no negative words *(e.g. send them home)*. Although they do not contain explicitly hateful words, they are conveying the desire to distance different groups, within which there is an inherent promotion of discrimination and division within society, fostering widespread societal tensions.

There have been a number of attempts to automatically identify and quantify cyberhate by using different approaches, such as lexicons, syntactic and semantic features - yet the limitation lies in classifying text that does not contain clearly hateful words and would have an impact on classification accuracy, *(e.g. send them home)* . While previous studies highlight the utility of methods capable of measuring semantic distances between words, such as embedding learning using individual words [15], and n-grams [35], this example requires an additional layer of qualitative context that sits above combinations of individual words.

Recent studies have begun to interpret the effective features for machine classification of abu- sive language by focusing on how language is used to convey hateful or antagonistic sentiment. 'Othering' - the use of language to express divisive opinions between the in-group ('us') and the out-group ('them') - has been identified as an effective feature [10]. The concept of 'othering' offers a potential candidate framework for the aforementioned qualitative layer capable of capturing the more subtle expressions of cyberhate such as the 'send them home' example. Anti-Hispanic speech might make a reference to border crossing or crime, anti-African American speech often references unemployment or single parent upbringing, and anti-Semitic language often refers to money, banking and the media. The use of stereotypes also means that certain types of language may be regarded as hateful even if no single word in the passage is hateful by itself [16]. In this study we develop a novel method for cyberhate classification based around (i) the use of two-sided pronouns that combine the in-group and out-group (e.g. your/our, you/us, they/we), and (ii) the use of pronoun patterns, such as verb-pronoun combinations, which capture the context in which two-sided pronouns are used (e.g. send/them, protect/us). Our hypothesis is that considering these linguistic features will provide an additional set of qualitative features that will improve the clas- sification performance. We use these to build a lexicon to develop advanced semantic similarity detection methods in the form of paragraph embeddings.

Paragraph embedding algorithms learn the semantic similarity of our proposed contextual features jointly with the rest of the text in the corpus. Samples that contain two-sided pronouns or pronoun patterns (e.g. verb-pronoun combinations) in a negative context are aligned in similar feature 'spaces'. This increases the probability of the machine classification method labelling any new samples exhibiting these features as cyberhate. For example, the following sentence: (**We** *want to hang* **them** *all*) contains the verb *"hang"* and pronoun *them*, as well as the two-sided pronouns *we, them*. If our hypothesis is correct, and such features do indeed improve the context of the automated learning method, we would expect the sentence: ***We*** *need to get* ***them*** *out* to be classified as cyberhate. This is not a sentence that would immediately flag as hateful by using existing classification methods, but is an example of where human annotators identified a threat to individual groups and communities, and therefore needs to be considered when 'taking the social mood' following trigger events. Our work builds on existing research by improving classification accuracy for four different types of cyberhate - religion, disability, race and sexual orientation - with F-measure improvements of: 8%, 11%, 11% and 4% respectively over state of the art research.

The paper is structured as follows: in Section 2 we review related work that is relevant to our technical proposal. In Section 3, we present our methods and explain the experimental steps. In Section 4, the classification results are presented and discussed. Finally, in Section 5, the contributions of this paper are summarized and some future directions are suggested for advancing this research area.

## 2 LITERATURE REVIEW

Here, we review academic work on cyberhate detection. This section is composed of four sub- sections: othering language narrative surrounding cyberhate, lexicon use for cyberhate detection, linguistic features for cyberhate detection, and the use of embedding learning.

### 2.1 Othering language

The hypothesis of this study is that we can leverage linguistic features of text posted to the Web to improve the classification of cyberhate. In particular we build a theoretical framework based on leveraging the theories of *othering* and *Intergroup Threat Theory (ITT)*. ITT posits that prejudice is a product of perceived realistic and symbolic threats. Realistic threats can be conceptualized in economic, physical and political terms. Such threats refer to competition over material economic group interests, including scarce resources such as jobs, houses, benefits and healthcare. Symbolic threats are based on perceived group differences in values, norms and beliefs. Out-groups that have a different viewpoint can be seen as threatening the cultural identity of the in-group [43]. Studies show that perceived threats to in-group values by immigrants and minorities are related to more negative attitudes towards these groups. For instance, research using ITT has recently focused on the perception of threat from Muslims in Europe [13]. This can result in 'othering' language, such as 'get them out', which represents a speech act that aims to protect resources for the in-group. The core concept is that these resources and values are threatened by the out-group, leading to anxiety and uncertainty in the in-group [44]. The desire to protect the in-group is considered the underlying motivation responsible for negative attitudes and discriminatory behavior. *Othering* is an established construct in rhetorical narrative surrounding hate speech [31], and the 'we-they' dichotomy has previously been identified in racist discourse [56]. Othering has been used as a framework for analysing racist discourse from a qualitative perspective in previous work. For instance, [55] argued that while the 'self' or the concept of 'us' is constructed as an in-group identity, the 'other' or the concept of 'them' is constructed as an out-group identity [47]. Therefore, polarization and opposition are created by emphasizing the differences between 'us' and 'them'. This may occur, for example, through the use of language to convey positive self-representation and negative representation of the 'other' as an out-group that is undesirable [54]. In machine learning research the principle of othering has been identified by [8] as a useful feature for classifying cyberhate based on religious beliefs, specifically for identifying anti-Muslim sentiment. However, this was post-classification in the form of an effort to interpret some of the statistically effective linguistic features. It is yet to be used as a theoretical foundation of feature engineering and tested with machine classification algorithms. Therefore the literature review of related work focuses on the classification performance of existing lexicon-based and linguistic features, as well as machine classification methods, to provide a baseline for comparison to our proposed innovation.

## 2.2 Lexicon-based Hate Speech Detection

While the dictionary-based approaches generally suffer from an inability to find offensive words with domain and context-specific orientations [14], corpus-based approaches use a domain corpus to capture opinion words with preferred syntactic or co-occurrence patterns. Focusing on a theme-related lexicon, [21] generated a lexicon of sentiment expressions using semantic and subjectivity features with an orientation towards hate speech, and then used these features to create a classifier for cyberhate detection. Work conducted by [53] proposed a method for automatic cyberhate detection using two steps: first, they used word features (tokens), sentence/structure features (dependency relations) and document features (document topic). They applied a binary feature that captured whether the word was being locally negated. Its value was true if a negation word or phrase was found within the four proceeding words, or in any of the word's children in the dependency tree. If there was no negation word in a phrase, this intensified the hateful rating. Secondly, they combined sentence-level features with dictionary-based features and achieved significant improvements in cyberhate polarity prediction by 4.3% compared to the baselines. Using pattern analysis [42] detected cyberhate using sentence structure - specifically patterns starting with the word 'I'. They assume that the word 'I' means that the user is talking about the emotions that he or she is feeling. Phrase-level sentiment analysis is different from lexicon-based analysis in that the former focuses on the polarity of the contextual content (e.g. whether there is negation in the sentence or not), whereas the latter predicts the polarity depending on the occurrence of the words that exist in the dictionary. [3] introduced a sentiment scoring method using emoticons, modifiers, negations and domain specific words. Lexicon methods could involve the use of offensive words and slurs or negative/positive related words (e.g. emotions and negation words) as features which might help to distinguish hate speech from other posts yet still have a weakness in detecting the hate stereotypes when the text contains no single hateful words, which will limit its effectiveness in detecting 'othering' language.

## 2.3 Linguistic Features

One of the most basic forms of natural language processing is the Bag of Words (BoW) feature extraction approach. BoW has been successfully applied as a feature extraction method for automated detection of hate speech, relying largely on keywords relating to offence and antagonism [9, 36, 50]. However, the method suffers from a high rate of false positives, since the presence of hateful words can lead to the misclassification of tweets being hateful when they are used in a different context (e.g. 'black') [22]. For instance, [14] demonstrated how non-hateful content might be misclassified due to the fact that it contains words used in racist text. In contrast, they also showed that hateful instances were misclassified because they did not contain any of the terms most strongly associated with cyberhate.

N-grams are features that capture consecutive words of varying sizes (from *1...n*) and have been used to improve the performance of hate speech classification by capturing context within a sentence that is lost in the BoW model [6, 11, 20, 35, 48]. [48] found that character n-grams have been shown to be appropriate for abusive language tasks due to their ability to capture variations of words associated with hate. In addition, [48], [32] found that using a character n-gram based approach outperforms word n-grams due to character n-gram matrices being far less sparse than the word n-gram matrices. [48] considered extra-linguistic features including gender and location. [30] aimed to detect cyberhate and distinguish it from general profanity by building surface n-grams, word skip-grams, and Brown clusters features for discriminating between hate speech and profanity using an annotated data set with three labels: (1) hate speech; (2) offensive language but no hate speech ; and (3) no offensive content. They found that a character 4-gram model was able to distinguish between hate speech and offensive language while other features were more frequently confused for offensive content. A study published by [39] used self-identified hateful communities as training data for hateful speech classifiers using a sparse representation of unigrams with tf-idf weights as a feature set. They clarify that the self-identified hateful communities outperforms the use of keyword-based methods by 10%-20%. In contrast, [18] showed the strength of leveraging the insult key word for capturing both explicit and implicit hate speech from an unbiased corpus. They trained two weakly supervised bootstrapping (slur term learner and LSTM classifier) models simultaneously to identify hateful tweets. They found that training two models jointly identified many more hate speech texts with a significantly higher (albeit low compared to other literature) F-score (slur learner = 0.19, LSTM = 0.26, both = 0.49). [12] examined a wide range of functionally- significant grammatical features (e.g. markers, place and time adverbials,questions, nominal forms, passives, adverbs, verbs..etc) to identify the main dimensions of functional linguistic variation that occur in racist and sexist Tweets. They identified 3 dimensions of linguistic variation in racist and sexist Tweets: interactive, antagonistic, and attitudinal. By applying different linguistic features, they demonstrated that there is a significant functional difference between racist and sexist Tweets, with sexists Tweets tending to be more interactive and attitudinal than racist Tweets. A study by [5] examined the paraphrasing of 35 tweets of implicit cyberhate. They identified tokens that are clear indicators for hateful content by retrieving words that are most strongly related with cyberhate by using nouns, named entities, and hashtags. They indicated that all (except one) of the significantly differing instances are perceived as more hateful in the implicit version, yet the paraphrasing method did not significantly advance the machine classifier results. [49] examined four feature extraction methods, namely: sentiment-based (specific method for scoring), semantic (e.g. punctuation, capitalized words, etc), unigram and pattern approach individually and then the four features were combined producing a fifth feature set. The pattern feature set was prepared using PoS tagging as follows: when the tweet contained sentiment, a specific PoS tagged word (e.g. coward) was replaced by *Negative_ADJECTIVE*, otherwise when the tweet did not contain any sentiment words they were replaced by a simplified PoS tag. They showed that the unigram features as well as the pattern features present the highest accuracy with values respectively equal to 82.1% and 70%, whereas the semantic and sentiment features did not produce a good classification accuracy. However, the combination of all previous features achieve accuracy equal to 87% for binary classification.

In general, while previous studies addressed the difficulty of the definition of hateful language, their experiments led to better results when combining a large set of features. They showed that BoW, n-grams, part-of-speech tagging, and data preprocessing (stop word/punctuation removal) provided a significant improvement in sentiment classification among different data sets (blogs and movies) when applied as a sophisticated combination of feature sets. They also speculated that engineering features based on deeper linguistic representations (e.g. dependencies and parse tree) may improve classification results for contents on social media. Typed dependencies have been widely used for extracting the functional role of context words for sentiment classification [23, 25] and document polarity [45]. [9, 10] demonstrated the effectiveness of applying typed dependencies for classifying cyberhate. Their study showed that typed dependencies consistently improved the performance of machine classification for different types of cyberhate by reducing the false negatives by 7%, beyond the use of BoW and known hateful terms.

## 2.4 Text Embedding

Embedding learning is aimed at training a model that can automatically transform a sentence/word into a vector that encodes its semantic meaning. It has been shown that embedding representation is very capable of semantic learning when word vectors are mapped into a vector space, such that distributed representations of sentences and documents with semantically similar words have similar vector representations [33] [34]. Based on the distributional representation of the text, many methods of deriving word representations that are related to cyberhate and offensive language detection are explored in the following works. In general, neural network applications were shown to be capable of capturing specific semantic features from complex natural language (e.g. location [37], entity [41] and images feature [2]). For hate speech detection purposes, [15] solved the problem of high dimensionality and sparsity by applying sentence embedding (paragraph2vec). In their study, paragraph2vec, which is an extended version of Word2Vec for sentences, has been shown to outperform the BoW representation for cyberhate classification models by around 3% to 4% in F1 score. However, they limited their study by comparing the classification results with TF-BoW and TF-IDF-BoW for the same comments. Similarly, [4] compared the classification accuracy of the combination of different baselines and classifiers (Char n-gram, TF-IDF, BoW and LSTM) and found that learning embedding with gradient-boosted decision trees led to the best classification performance by 18% over state-of-the-art char/word n-gram methods. For the German language, [40] examined different types of features (BOW, 2-grams,3-grams, linguistics, Wod2Vec, Doc2Vec, extended 2-grams and extended 3-grams) for training logistic regression LR classifiers. The experimental results, obtained on a 75/25 split between training and test data, showed that the best performing types of features are Word2Vec and Extended 2-grams.

Word vector extraction was also applied to tweets for cyberhate classification by [17], who built a Convolutional Neural Network (CNN) model which was trained on four feature sets: character 4-grams, word vectors based on semantic information built using Word2Vec, randomly generated word vectors, and word vectors combined with character n-grams. They showed that while adding character n-grams slightly increased the precision but resulted in lower recall and F-measure; the second feature set (Word2Vec alone) performed best overall, with an F-measure of 0.78. Recently, [58] introduced a deep neural network model combining CNN and gated recurrent unit GRU layers, which were used to train on Word2Vec features. The results show that the classification accuracy was improved by between 1 and 13% when compared to baselines methods.

Several works have merged typed dependencies with embedding learning and clarified the different levels of embedding learning when using the dependency context rather than the bag of words linear text. [29] showed that dependency context embeddings can provide valuable syntactic information for sentence classification tasks, which is a motivation for implementing classification tasks in respect of dependency embedding text, and [26] showed that dependency- based embeddings are less topical and exhibit more functional similarity than bag of words linear embeddings . In addition, [59] defined the differences between flat text, which they called neighbour words, and the dependency context, and clarified through examples the drawbacks of learning embedding of flat text. While these studies introduced the effectiveness of the word distances of the dependency context, which capture the semantic relations, their works targeted other areas of research, not cyberhate. One study that *has* combined typed dependencies with embedding learning in the context of cyberhate was reported by [35]. They developed a machine learning approach to cyberhate based on different syntactic features as well as different types of embedding features, and reported its effectiveness when combined with some standard NLP features (n-grams, dependencies) in detecting hate speech in online user comments. They showed how applying each feature set alone resulted in different classification performance, and found that character n-grams alone are useful in noisy datasets. While using n-grams boosted training performance, n-grams result in high dimensionality and thus render the models susceptible to overfitting. They examined different syntactic features as well as different types of embeddings features and showed that this combination outperforms basic embedding learning. While they examined a combination of different syntactic features and embedding learning, some syntactic features led to confusion in the embedding learning process (e.g. some PoS and dependency modifiers). Therefore, this presents an open research question on how to better refine the framing of cyberhate from a computational feature processing perspective to improve the classification performance. Our study is different from the previous study in that we are yet to see evidence that the complex and nuanced 'us and them' narrative emerging on social media can be captured using a combination of typed dependencies and embeddings.

## 3 THEORY AND DATA

In this section we introduce our research hypothesis, the datasets that used in our study, and a statistical analysis on the use of othering language in our datasets.

### 3.1 Research Hypothesis

Based on the analysis of existing literature on automated approaches to cyberhate and the theories of Integrated Threat Theory and 'othering' as a form of discriminatory language use, we propose that existing machine classification performance could be improved by including a layer of linguistic features representing 'othering' within short informal text, such as posts that are published to online social networks. We assume that othering language could be identified through specific uses of English language parts of speech - particularly verbs (action-driven language) and pronouns (referring to 'others'). We hypothesize that the use of pronouns that refer to an ingroup (e.g. we, us) *co-occurring* with pronouns that refer to an outgroup (e.g. them, they) in the same post, will be indicative of divisive or antagonistic attitudes and therefore will improve machine classification of cyberhate. In this study, we refer to the co-occurrence of ingroup/outgroup pronouns as a *two-sided pronoun*.

Figure 1 presents an overview of linguistic features that can be used between different groups to distinguish themselves (the in group) from others (the out group). The figure shows how pronoun terms from one side (us, we, our, etc.) draw the boundary between the in group by referring to the the out group (we, they etc).

### 3.2 Datasets

We used two datasets for our experimentation. Our main contribution to the literature is to enhance existing cyberhate machine classification performance by developing a novel othering lexicon to provide additional features that capture more nuanced forms of cyberhate based around a fear of 'the other' - for instance, around immigration and terrorism. To develop the othering lexicon we used the dataset provided by [14]. They collected tweets containing different types of hate and used crowd-sourcing to further divide the sample into three categories: those containing hate speech, those with only offensive language, and those with neither. We refer to this as *Dataset 1*. To compare our work to the state-of-the-art in cyberhate classification we used a second dataset, which was used in previous work [10]. The dataset was collected from Twitter and annotated using the CrowdFlower human intelligence task service with a single question: 'Is this text antagonistic or hateful based on a protected characteristic?'. The dataset comprises cyberhate directed at four different protected characteristics, as follows: sexual orientation - 1803 tweets, with 183 instances of offensive or antagonistic content (10.15% of the annotated sample); race 1876 tweets, with 73 instances of offensive or antagonistic content (3.73% of the annotated sample); disability 1914

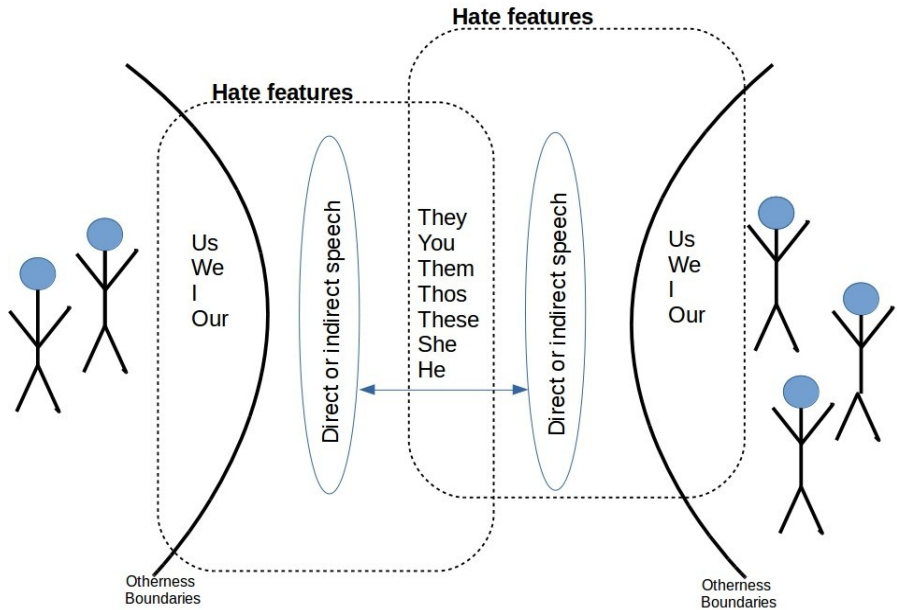

Fig. 1. The graph shows the boundary defined between two groups using the othering terms, and the space between the boundary shows how the negative text could be defined

tweets, with 51 instances of offensive or antagonistic content (2.66% of the annotated sample); and religion 1901 tweets, with 222 instances of offensive or antagonistic content (11.68% of the annotated sample). The authors conducted all of the necessary tests so as to ensure agreement between annotators for the gold-standard samples [10]. The amount of abusive or hate instances is small relative to the size of the sample. However, these are random instances of the full datasets for each event and they are considered representative of the overall levels of cyberhate within the corpus of tweets [9]. We evaluate the relative improvement in classification performance using this dataset, which we refer to as *Dataset 2*.

### 3.3 Summary Statistics for Othering Language in the Datasets

To provide an initial justification for our research hypothesis on the use of two sided pronouns to improve cyberhate classification, we conducted a corpus analysis of our datasets - calculating the percentage of tweets from Dataset 2 that included at least two pronouns. We found these features present in only 0.9% of non-malicious instances within the data and 17.6% of cyberhate instances. In Figure 2 we show the comparative occurrence of two-sided othering terms in both hateful samples and non-hateful samples among different types of cyberhate datasets. We can see the anti-Muslim dataset contains the most frequent usage of the two-sided othering language, followed by anti-semitism. This follows from the findings of [8] who identified 'othering' language was a useful feature for classifying cyberhate based on religious beliefs. We can also note that the percentage of two sided othering was higher in the hateful annotated samples than non-hateful samples for all other type of cyberhate, and to around the same level.

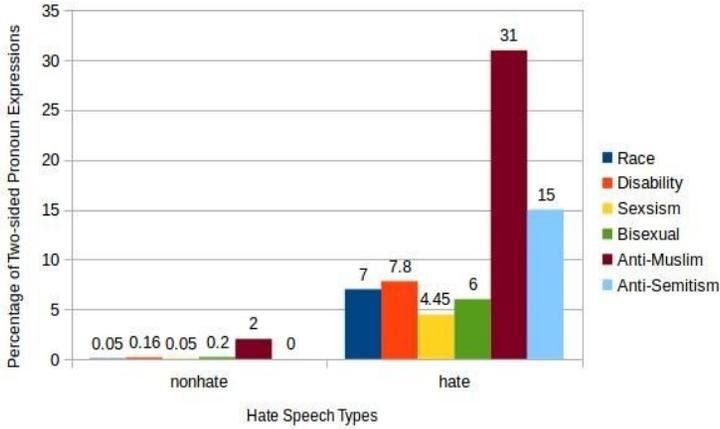

Fig. 2. The graph shows the use of two-sided othering for each hate speech type in both hateful and non- hateful samples

Clearly, this strategy will not identify all existing hate speech in social media but we propose that this additional contextual layer will improve the performance above that of existing literature.

## 4 METHODS

To achieve our novel othering layer within the machine classification framework we developed a lexicon containing three components: (i) a constrained subset of dependency relationship labels extracted using probabilistic parse trees that we hypothesized would be representative of othering, (ii) more general parts of speech associated with othering including verbs (VB), nouns (NN) and adjectives (JJ) (e.g. send them home), and (iii) a list of English pronouns. Together these capture as much as possible linguistically to provide a focused set of othering features. This section details the process used to extract these features and how they were used in our automated machine classification approach.

### 4.1 Extracting Othering Terms

The first phase involved using Dataset 1 and analysing only the hateful samples. We identified all samples where two-sided othering was present - i.e. where at least two pronouns were used. We discarded any samples without at least two pronouns. Using this subsample, we used the Stanford Typed Dependency Parser to transform the text to provide co-occurring words with a probabilistically derived linguistic label. Figure 3 shows the linguistic labels associated with each word in a sample sentence.

The Stanford Parser returns 51 different linguistic labels, but to provide a specific focus on other- ing language, we retained only 6 types of dependency relationships: *nsubj*, *dobj*, *nmod*, *det* , *advmod* and *compound*. The remaining dependency modifiers were discarded. The rationale for preserving these modifiers is as follows. The *nsubj* label captures the syntactic subject or proto-agent in a sen- tence (i.e. the active agent). Examples include 'muslims caused' and 'they inflicted'. *dobj* concerns the direct object of a verb phrase and has a high probabilistic likelihood for capturing relationships between verbs and nouns, pronouns and determiners in the same phrase (e.g. send and them). *nmod* is likely to identify nominal modifiers for nouns, for instance 'all gays' or 'womens place'. The *det* captures the relationship between nominals and their determiner (e.g. 'these terrorists'). *advmod* captures adverb modifiers (e.g. where we see 'home' we may also see 'send'). The *compound* will identify compound verb phrases including verb and adjective compounds such as 'send back' or 'kill black'.

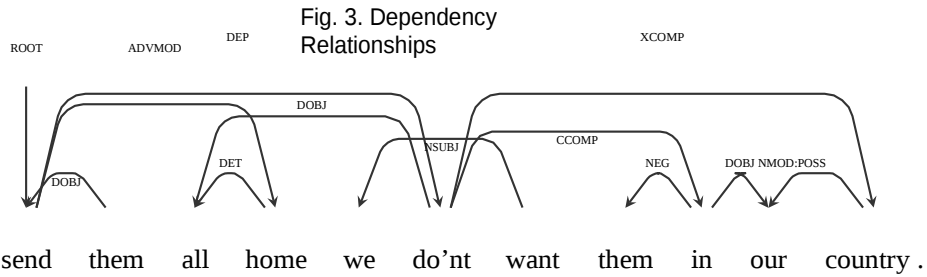

Fig. 3. Dependency Relationships

As a worked example of how this method is expected to capture othering, the translation of the text in Figure 3 becomes *nsubj(want-7, we-5), dobj(send-1, them-2) det(home-3, all-4), nmod:poss(country- 11, our-10)* and the remaining relationships would be discarded. We are now capturing distinctive othering features that co-occur in the same sentence. Despite none of these words being clearly antagonistic or offensive of their own - together they provide a greater contextual feature for machine classification to detect this is unseen samples using similar phrasing.

### 4.2 Building the Othering Lexicon

To compliment the dependency relationship features we also applied part-of-speech (POS) tagging to the hateful samples that included at least two pronouns. We once again refined the set of labels to include those most likely to represent othering and retained only words tagged as nouns (NN), adjectives (JJ), verbs (VB) and adverbs (RB). The POS labels themselves were removed to leave only words. These POS words, the dependency relationship features and a list of all English pronouns were then concatenated into a triple that formed the basis of an othering lexicon - a novel concatenation of a range of grammatical and linguistic features extracted from a human annotated data set of hateful and antagonistic texts. The complete workflow for othering feature extraction is shown in top part of Figure 4.

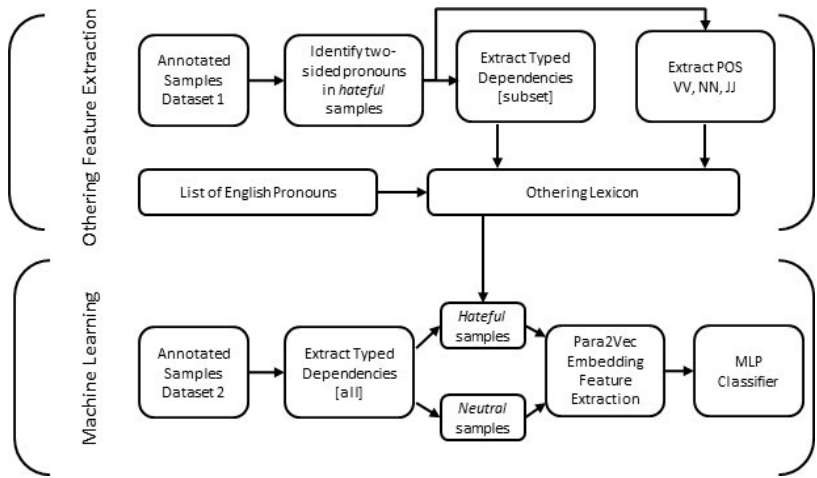

Fig. 4. Othering Feature Extraction and Machine Classification Workflow

### 4.3 Embedding Learning

At this stage, we need to identify a suitable method for utilising the features extracted through our othering lexicon. We could have used these as raw features for classification but to provide further refinement we employed embedding learning to capture the relative 'distance' between these features in a cyberhate context. Learning vector representations allows us to plot each feature in such a way that we can calculate numeric distances between features based on their use in a context. A common example of this is that the distances between 'man' and 'king' would be similar to that of 'woman' and 'queen'. Therefore we can identify relationships between words (i.e. 'man' and 'king'), and context (i.e. 'king' is to 'queen' as 'man' is to 'woman'). With two-sided pronouns we assume that our method would benefit from this approach to extract the semantic 'meaning' of othering features and learn these jointly across the hateful and non-hateful texts to provide context for term use - ultimately with the aim of these features being able to better support machine classification of both.

Various methods have been proposed to learn vector embedding representations. Word2Vec and Doc2Vec have been proposed for building word/paragraph representations in low-dimensional vector space [33]. In the Word2Vec model, *words* are represented in continuous space where semantically similar words have a high similarity measure in that space. In the Doc2Vec model, *paragraphs* are represented as low-dimensional vectors and are jointly learned with distributed vector representations of tokens using a distributed memory model (for further detail see [34]). Every sentence is mapped to a unique vector, and every word included in the sentence is also mapped to a unique vector. In our context this means each tweet is fed into the embedding learning methods as if it were a sentence, and each feature of the tweet derived in the othering feature extraction phase (see top of Figure 4) becomes a word. The words of the tweet are represented at one layer of vector embedding abstraction, while the othering features are learned at another layer. *This is what provides us with our key contribution in this paper - the fact we have added another layer of abstraction to the learning process that is specifically focused on othering features.*

As shown at the bottom of Figure 4 - prior to implementing the embedding phase we introduced *Dataset 2*, which includes the labelled data from previous studies [9][10] for evaluation purposes.

Dataset 2 was parsed using the Stanford Typed Dependency parser. At this point the Typed Dependency features from the hateful samples in Dataset 2 were concatenated with our othering lexicon. The transformed hateful and non-hateful samples were then fed to the Doc2Vec algorithm and the embedding were computed jointly across both sets samples.

We learned the Distributed Memory (PV-DM) vectors using the Gensim [1] implementation of distributed representations of the sentence (tweet) [27]. In the Distributed Memory component (PV-DM), the sentence acts as a memory that remembers the missed word in the current context of the sentence. According to [34], the distributed memory model is consistently better than PV-DBOW. To find the best implementation for our data, we experimented with both and found that distributed memory performed better in learning our data set vectors. We used small window sizes because, according to [29], a window of size 5 is commonly used to capture broad topical contents, whereas smaller windows (e.g. *k=2* windows) contain more focused information regarding the target word.

For example, for $k = 2$, the context around the target word *w* comprises $w-2, w-1, w+1, w+2$. These become the features used for learning distances between the target word and its surrounding context. The larger the window, the broader the context. The final output is a vectorised data set that is used as a feature set for feeding in to a machine classification approach. We were expecting the othering layer to assist in improving the performance of machine classification of cyberhate, so a more focused approach seemed logical given it will be these nuanced othering terms in a smaller window that will likely lead to improvements in classification. We experimented with various window sizes including 100, 300, 600, 800 and 1000 dimensions and $k = 2, 3, 5, 6$ and 10. We recorded the performance of each and report the best performing configuration - which was for 600 dimensions and *windows = 2*.

### 4.4 Machine Classification

Once we learned vector representations, we joined the vector with its original human-assigned label, assigning the label 0 to the non-hateful samples and 1 to the hateful samples, and then used these to train and test the machine classifier. We examined several classifiers that have been used in the related work to determine the overall improvement when using these classifiers with our novel othering feature lexicon. The candidate methods included Support Vector Machines (SVM), Decision Trees (DT), Naive Bayes (NB) and Random Forests (RF), as used in [9][10] and Logistic Regression (LR) as used by [15]. The SVM parameters were set to normalize data, use a gamma of
0.1 and C of 1.0 and we employed radial basis function (RBF) kernel. Decision tree classifiers (DT) iteratively identify the feature from the vector that maximizes information gain in a classification, and the Random Forest (RF) iteratively selects a random sub-sample of features in the training stage and trains multiple decision trees before predicting the outputs and averaging the results which maximize the reduction in classification error [7]. Random Forest Decision Tree algorithm was trained with 100 trees. In addition to the previous classifiers, we also examined the Multilayer Perceptron (MLP) classifier [57]. MLP is a feed-forward artificial neural network model which maps input data sets on an appropriate set of outputs. MLP consists of multiple layers of nodes in a directed graph, with each layer being fully connected to the next layer [19]. [24] demonstrate that multilayer feed-forward networks can provide competitive results on sentiment classification and factoid question answering. We examined MLP with two hidden layer, five hidden connected units, and 200 iterations.

---
[1] https://radimrehurek.com/gensim/models/Doc2Vec.html

# 5 RESULTS AND DISCUSSION

The baseline results for this work stem from the previous state-of-the-art, produced in [10], [15] and [35]. The previous studies use different feature sets for cyberhate classification, part of which we applied in our study. [10] used a Bag of Words (BoW), n-gram, and Typed Dependency features. [15] used Doc2Vec for joint modeling of comments and words and [35] combined different NLP features with Doc2Vec and Word2Vec, - as we have in our study, but upon which we have developed a novel layer of features to capture othering. They used a Logistic Regression model for classification using these features. We implemented all these approaches and have provided results for comparison with our proposed othering lexicon. Throughout the results section we refer to the correct classification of non-hateful samples as true negatives, and correct classification of hateful samples as true positives.

## 5.1 Quantitative Results

We trained and tested our approach as well as the baseline methods using ten-fold cross validation. This method has previously been used for experimentally testing machine classifiers for short text [46] [10]. It functions by iteratively training the classifier with features from 90 percent of the annotated data set, and classifying the remaining 10 percent as 'unseen' data, based on the features evident in the cases it has encountered in the training data. It then determines the accuracy of the classification process and moves on to the next iteration, finally calculating the overall accuracy. The results presented in Tables 1 and 2 are for the cyberhate class only. The classification performance for non-hateful text was consistently above 0.90-0.99 and are omitted to reduce complexity in presenting the results. Our main interest is with the improvement of cyberhate classification.

We use F-measure as our main comparison metric, given it controls for false positives and false negatives and a lack of balance in the dataset. Our first set of experiments were conducted to test the performance of the range of machine classification methods proposed in Section 4.4. The features used included those extracted in the feature preparation stage detailed in Section 4.1-4.3 - including the core contribution of this research - the othering lexicon. Table 1 shows the results of the range of classifiers, with the Multilayer Perceptron neural network architecture performing best in 3 out of 4 cases. We propose that this model be used in the comparative experiments against the baseline methods on this basis, especially given the MLP model appears to find the best balance between false positive (FP) and false negative (FN) classifier outputs. The Logistic Regression model used by [15] comes close to MLP, and performs best in the disability class - but even then there is a disproportionate balance of FP and FN (0 FP and 11 FN), whereas MLP has a lower F-measure in this case but a better balance of FP and FN (2 FP and 10 FN).

Table 1. F-score measurements of Different Classifiers Performance on our Framework

| | | Religion | Disability | Race | Sexual-Orientation |
|---|---|---|---|---|---|
| Othering lexicon + Doc2Vec | SVM | 0.72<br>FP=0, FN=96 | 0.85<br>FP=0, FN=13 | 0.81<br>FP=0, FN=22 | 0.51<br>FP=5, FN=121 |
| | NB | 0.86<br>FP=58, FN=10 | 0.85<br>FP=6, FN=9 | 0.81<br>FP=24, FN=6 | 0.90<br>FP=21, FN=15 |
| | LR | 0.91<br>FP=6, FN=30 | 0.88<br>FP=0, FN=11 | 0.91<br>FP=0, FN=12 | 0.95<br>FP=2, FN=16 |
| | DT | 0.90<br>FP=23, FN-22 | 0.87<br>FP=8, FN=6 | 0.93<br>FP=3, FN=6 | 0.88<br>FP=33, FN=15 |
| | RF | 0.92<br>FP=13, FN=19 | 0.84<br>FP=3, FN=11 | 0.96<br>FP=0, FN=5 | 0.95<br>FP=5, FN=12 |
| | MLP | 0.93<br>FP=4, FN=19 | 0.86<br>FP=10, FN=2 | 0.97<br>FP=1, FN=4 | 0.98<br>FP=6, FN=2 |

Our second set of experiments involved comparing our model to those of the baseline results using methods published in the state of the art research. We conducted two experiments of our own. One using the othering lexicon alone (Proposed model 1), and one using the othering lexicon with Doc2Vec embeddings (Proposed model 2). The reason for performing these separately was to demonstrate the added value of using the othering lexicon not simply as an additional 'Bag of Words' - i.e. where each feature in the lexicon was used individually, but as a set of features from which we could extract a form of semantic meaning through computationally determining relative distances between terms. The detailed results of these experiments are shown in Table 2, and Figure 5 shows the results as a comparison of F-measure for each model.

At an individual type level, the baseline results for religion were 0.77, 0.83 and 0.85 for [10] (Baseline model 1), [15] (Baseline model 2) and [35] (Baseline model 3) respectively. Using the othering lexicon alone we were able to show an improvement to 0.90. This was mainly due to a large reduction in false negatives (missed instances of cyberhate) - most likely due to lack of dependency on individual words while picking up the 'us and them' narrative. When incorporating the embedding approach as well, we saw improvements up to 0.93 F-measure. For disability the baselines results were 0.75, 0.11 and 0.43. This type of cyberhate has caused problems for machine classifiers in previous research as it contains very few clearly hateful sentiments. Generally the narrative is more along the lines of mockery and jest. Indeed using our othering framework alone we were unable to perform well on this type, achieving a results of 0.55. However, the embedding method was able to achieve a result of 0.86 - showing this approach can also detect slightly different forms of othering based around mockery. For race the baseline results were 0.75, 0.24 and 0.86. The othering approach alone scored only 0.77 F-measure, which is below that of baseline model 3. It has a higher number of false positives - more than double the number of the baseline model. However, the inclusion of the othering lexicon with the embeddings method boosted this to 0.94 with only 4 FP and 1 FN obtained. This was the best performing model across all types and arguably stands the most chance of exhibiting othering where ingroup and outgroup are defined by skin colour. Finally, for sexual orientation, the baseline results were 0.47, 0.94 and 0.91. Again we see a dip in performance in comparison to baseline models 2 and 3 when using the othering lexicon alone. It produced no false negative, but significantly higher false positive. Though as for race, we see an improvement when combining the othering lexicon with embeddings, creating the additional abstracted layer of othering and improving the result to 0.98 F-measure.

Table 2. Machine classification performance for cyber hate based on Religion, disability, race and sexual orientation

| Features | classifier | Religion | | | Disability | | | Race | | | Sexual-Orientation | | |
|---|---|---|---|---|---|---|---|---|---|---|---|---|---|
| | | P | R | F | P | R | F | P | R | F | P | R | F |
| Baseline model 1: n-Gram words (1-5) with 2,000 features +n-Gram typed dependencies +hateful terms [10] | SVM | 0.89 FP=19 | 0.69 FN=70 | 0.77 | 0.97 FP= 1 | 0.61 FP= 20 | 0.75 | 0.87 FP= 7 | 0.66 FN= 24 | 0.75 | 0.72 FP= 25 | 0.35 FN=119 | 0.47 |
| Baseline model 2: Comment Embedding [15] | MLP | 0.79 FP=47 | 0.88 FN=22 | 0.83 | 0.18 FP=47 | 0.08 FN=18 | 0.11 | 0.16 FP=11 | 0.50 FN=59 | 0.24 | 0.91 FP=14 | 0.95 FN=4 | 0.94 |
| Baseline model 3: N-grams+ linguistic+ dependencies+ word and comment Embedding [35] | MLP | 0.84 FP=35 | 0.86 FN=29 | 0.85 | 0.29 FP=36 | 0.83 FN=30 | 0.43 | 0.85 FP=10 | 0.88 FN=8 | 0.86 | 0.88 FP=22 | 0.95 FN=8 | 0.91 |
| Proposed model 1: The othering lexicon | MLP | 0.82 FP=40 | 0.82 FN=1 | 0.90 | 0.78 FP=29 | 0.43 FN=6 | 0.55 | 0.64 FP=24 | 0.93 FN=3 | 0.77 | 0.80 FP=36 | 1.00 FN=0 | 0.88 |
| Proposed model 2: Othering lexicon + Paragraph2vec Embedding | MLP | **0.98** FP=4 | **0.89** FN=19 | **0.93** | **0.80** FP=10 | **0.95** FN=2 | **0.86** | **0.94** FP=4 | **0.98** FN=1 | **0.97** | **0.97** FP=6 | **0.99** FN=2 | **0.98** |

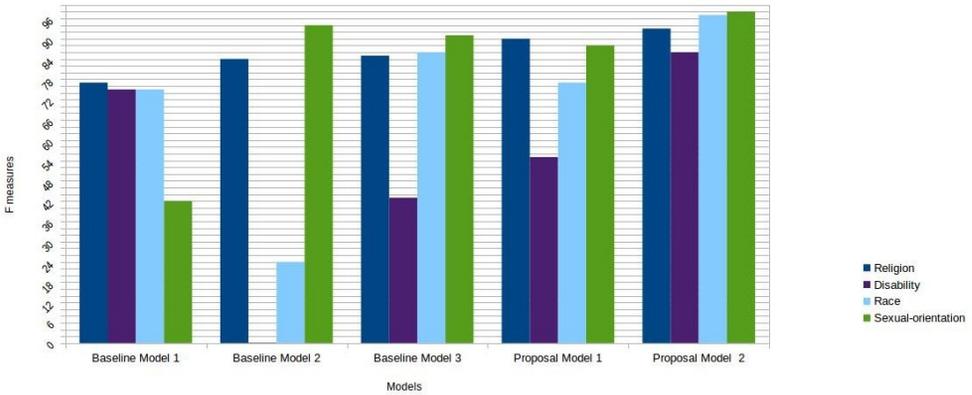

Fig. 5. F-scores for each model

The key finding from previous research was that the inclusion of features capable of detecting othering language in the classification of religious cyberhate reduced the false negative rate by 7%, when comparing with using hateful terms alone. In addition, [35] found that computing feature embeddings when combining with the standard NLP features showed promise to improve the performance for cyberhate classification. In our experiments, the results have shown how the use of othering features alongside embeddings has enabled us to present a new level of abstraction to the classifier in the form of othering-level feature embeddings. For each type of cyberhate, we see the following F-measure improvements over the state of the art of the best performing: religion sees a 8% improvement (over baseline model 3), disability is improved by 11% (over baseline model 1), race by 11% (over baseline model 3) and sexual orientation by 4% (over baseline model 2).

Furthermore, the general finding from our results is that the inclusion of the othering lexicon in the semantic learning embedding phase improves the classification results to the point that our proposed approach outperforms the state of the art baseline methods in *all four types of cyberhate*. The previous best baseline results that come closest to ours are spread across all three baseline models - therefore our approach moves the field of cyberhate classification closer to a universal model of cyberhate.

## 5.2 Qualitative Results

Given our improvements over the state of the art using the othering lexicon, and the post-computation analysis conducted in [10] on the statistically significant features associated with the machine classifier labelling an unseen sample as hateful or antagonistic, we conducted our own qualitative analysis to identify any insights into the features captured using feature embedding on the othering features - i.e. the two-sided pronouns. Given the improvement, we can assume that the embedding method has effectively assigned othering features to similar vectors spaces in such a way as to better distinguish hateful from non-hateful content using the Doc2Vec embedding algorithm. We have visualized our model in two ways - first using embeddings only - which would lead to the features seen in the baseline models (see Figure 7), and second using the othering lexicon with the embedding model (see Figure 6). We visualise our model using TensorBoard which has a built-in visualizer (we perform 2D principal component analysis(PCA)), called the Embedding Projector, for interactive visualization and analysis of high-dimensional data like embeddings [2]. We used the sample from the Religion type in *Dataset 2* for this. The distances between words are relative based on their computed similarity to other words in the hateful sample. The two graphs are focused and enlarged to show the 300 most similar words. The colors indicate the distances from the key word 'us', the purple dots indicate the smallest distances (0.008-0.09), next smallest are the pink dots (0.093-0.2), then oranges dots (0.21-0.39), then dark yellow dots (0.4-0.55), and finally the light yellow dots are furthest from 'us' (0.56-1>). In distance functions, smaller values imply greater similarity between words [1]. Ideally we want the classifier to be able to use these small distances to make effective use of them as features for distinguishing hate from non-hate.

We can see from Figure 6 that the words with the smallest relative distance from 'us' (the purple dots) include pronouns from the ingroup (*us, we etc.*), pronouns from the outgroup (*you, them, they etc.*), and different reaction verbs (*send, hang, stay, campaign etc.*), which captures their co-occurrence in more nuanced othering aspects of cyberhate language. From an Integrated Threat Theory perspective we can also see symbolic and realistic anxiety present (e.g. the words 'anxiety', 'campaign', 'confus(ion)' and invad(e)). We can also see the obvious derogatory terms in the same Figure (e.g. *raghead, wogs, nigger*). Thus, this model is picking up both the obvious and non-obvious cyberhate using our othering lexicon. Whereas in Figure 7 none of the words are particularly close to 'us' in terms of distance, meaning the classifier is unable to make effective use of the othering narrative. Thus, the classifier based on these features will become more dependent on the hateful words and miss the less obvious narrative. We posture that this ability to capture the more nuanced text is the core reason behind our successful improvement over the state of the art from previous research.

---

[2] https://www.tensorflow.org/versions/r0.12/how_tos/embedding_viz/

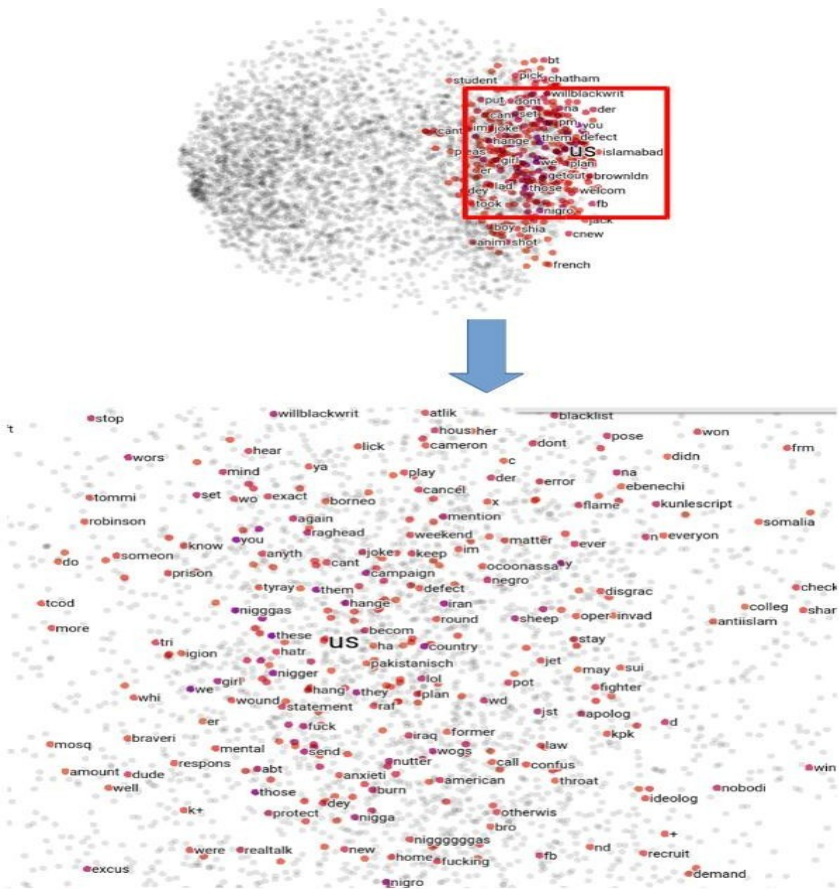

Fig. 6. Embedding and our Lexicon Visualisation on Religion data set

Fig. 7. Embedding Visualisation on original Religion data set

## 6 CONCLUSION

In this paper we aimed to improve the machine classification performance for different types of hateful and antagonistic language posted to Twitter - known as cyberhate. Our study was inspired by the concepts presented by Integrated Threat Theory (ITT) and 'othering' theory. We investigated the effectiveness of developing an abstract layer of linguistic features based around the use of 'othering' language, such as terms and phrases that separate the ingroup (e.g 'we', 'us') from the outgroup (e.g. 'them', 'those'), and suggest action or separation based on perceived symbolic and realistic threats (e.g. 'send them', 'get out'). Our hypothesis was that this additional layer would provide better context for the classifier beyond words alone. We used vector embedding and the Doc2Vec algorithm to cluster these features, thereby re-framing the linguistic features from individual terms and phrases to numeric distances representing a form of semantic similarity of these terms - learned in the context of hateful or non-hateful texts. We then experimented with machine classification methods to determine the improvement of our novel 'othering lexicon' over the state-of-the-art research, and the most effective machine classifier to use with our embedding-transformed feature set.

We tested the othering lexicon - comprised of a range of linguistic features using datasets from previous research that represented cyberhate targeted at specific social groups based on personal characteristics relating to religion, race, disability and sexual orientation. We found that our novel approach to othering feature identification provided improvements over three other leading approaches that were used as baseline models, such that religion saw an 8% improvement, disability was improved by 11%, race by 11%, and sexual orientation by 4%. Furthermore, our proposed approach outperforms the state of the art baseline methods in *all four types of cyberhate*. The previous best performing baseline results that come closest to our are spread across all three baseline models - therefore our approach moves the field of cyberhate classification closer to a universal model of cyberhate. Our results provide F-measure results for religion at 0.93, disability at 0.86. race at 0.97 and sexual orientation at 0.98.

We performed a qualitative inspection of the embedding representation of one type of cyberhate - religion - and were clearly able to see the vector space similarity between ingroup/outgroup terms (us, we, they, them, you), action terms (get, send, stay, hang), and terms related to anxiety of the threat from 'the other' (campaign, anxiety, confusion, invasion) when pre-processed using our othering lexicon, in comparison to the approach in the existing literature, which was to use embedding without the context of othering. This additional contextual layer provides the key novelty in our approach - it provides a word-level abstraction, learned at a different layer to the rest of the text, and the Doc2Vec embedding method has made good use of this additional layer to improve on the state of the art. Our approach allows the machine classifier to use a more broad range of features beyond individual words and ngrams, providing greater context for the classifier. In future we aim to develop larger datasets on which to test these classifiers and use them to study rising and falling cyberhate levels on a range of online social media platforms, with the intention of collecting these narratives to better understand the topics and touch-points that are being discussed in this context at an aggregate level during times of civil unrest or following trigger events.

## ACKNOWLEDGMENTS